\documentclass[runningheads]{llncs}
\usepackage{graphicx}    

\newcommand{\our}{\emph{L2T-FMT}}
\usepackage{booktabs}
\usepackage{multirow}
\usepackage{xcolor}
\usepackage[colorinlistoftodos]{todonotes}
\usepackage{subcaption}
\usepackage[linesnumbered,ruled,vlined]{algorithm2e}
\usepackage{algpseudocode}
\usepackage{amsmath}

\SetCommentSty{mycomm}
\usepackage{enumitem}
\usepackage{amssymb}

\SetKwInput{KwInit}{Init}       
\SetKwInput{KwInput}{Input}  
\SetKwInput{KwOutput}{Output}
\DeclareMathOperator*{\argmin}{argmin}
\DeclareMathOperator*{\argmax}{argmax}

\begin{document}

\title{Learning to Teach Fairness-aware Deep Multi-task Learning}

\author{Arjun Roy\inst{1,2}
       \and Eirini Ntoutsi\inst{2}
}
\institute{
              L3S Research Center, Leibniz University Hannover \and
              Institute of Computer Science, Free University Berlin 
              \email{\{arjun.roy,eirini.ntoutsi\}@fu-berlin.de}
              }





\maketitle
\begin{abstract}

Fairness-aware learning mainly focuses on single task learning (STL). The fairness implications of multi-task learning (MTL) have only recently been considered and a seminal approach has been proposed that considers the 
fairness-accuracy trade-off for each task and the performance trade-off among different tasks. Instead of a rigid fairness-accuracy trade-off formulation, we propose a flexible approach that learns how to be fair in a MTL setting by selecting which objective (accuracy or fairness) to optimize at each step. 
We introduce the \our~algorithm that is a teacher-student network trained collaboratively; the student learns to solve the fair MTL problem while the teacher instructs the student to learn from either accuracy or fairness, depending on what is harder to learn for each task. 
Moreover, this dynamic selection of which objective to use at each step for each task reduces the number of trade-off weights from 2$T$ to $T$, where $T$ is the number of tasks. Our experiments on three real datasets show that \our~improves on both fairness (12-19\%) and accuracy (up to 2\%) over state-of-the-art approaches. 

\end{abstract}
\section{Introduction}\label{sec.intro}



Multi-Task Learning (MTL)~\cite{2021mtl_survey} aims to leverage useful information contained in multiple tasks to help improve the generalization performance over all tasks. It is inspired by human's ability to learn multiple tasks and it has been already successfully applied in a variety of applications from natural language processing~\cite{ecml_mtl_2021vogue} to vision~\cite{ecml_mtl_2021self}. Many methods and deep neural network architectures~\cite{ecml_mtl_2021self,multi-task_gnn_2021deep,ecml_mtl_2021vogue} for MTL have been proposed, still the basic optimization problem is formulated as minimizing  a weighted sum of task-specific losses, where the weights are inter-task trade-off hyperparameters used to avoid 
inter-task loss dominance~\cite{2021mtl_survey}.



Despite its popularity, the fairness implications of MTL have only recently come into focus~\cite{MTLfairWang0BPCC21}. The area of fairness-aware learning for Single Task Learning (STL) has received a lot of attention in the last years~\cite{mehrabi2021survey} and methods have been proposed that aim to learn correct predictions without discriminating on the basis of some protected attribute, like gender or race.
Methods in this category reformulate the classification problem by explicitly incorporating the model's discrimination
behavior in the objective function through e.g., regularization or constraints.
The fair-MTL problem was only recently introduced and a solution, \emph{MTA-F}, has been proposed~\cite{MTLfairWang0BPCC21} that considers \emph{intra-task} fairness-accuracy trade-offs (as is typical in MTL) and \emph{inter-task} performance trade-offs (as is typical in fairness-aware STL). For $T$ tasks, such an approach requires $2T$ trade-off weights. The current practice to find the 
correct trade-off weights is by hyperparameter tuning~\cite{MTLfairWang0BPCC21} using common search techniques.  The complexity of such an approach can grow exponentially in time with each added task~\cite{2021hyperparameter}  and therefore does not scale well for a large number of tasks. 
Moreover, the trade-off weights in \emph{MTA-F} are fixed throughout the training process.
However, the
inter-task trade-offs 
and the intra-task fairness-accuracy balance may change over the training process due to e.g.,   external factors like the training batch~\cite{chen2018gradnorm}.
In this paper, instead of a fixed fairness-accuracy trade-off formulation, we propose to \emph{dynamically} select one among fairness and accuracy objectives at each training step for each task. To this end, we formulate the fair-\verb|MTL| problem as a student-teacher problem and propose the Learning to Teach Fair Multi-tasking (referred to as \our) algorithm. Our design inspiration comes from recent learning to teach (L2T) algorithms~\cite{fan2018learning2Teach,wu2018L2Tdynamic}. The student in our proposed algorithm is the desired \verb|MTL| model, which follows the instruction of the teacher to learn from the available accuracy or fairness objectives for each task, and updates its parameters accordingly. The student sends feedback about its progress on fairness and accuracy in each task to the teacher. The teacher learns from the feedback and updates its model. This way, both student and teacher networks are trained collaboratively. 
Except for the dynamic intra-task loss selection, we also propose to set the inter-task parameters \emph{dynamically} at each training step using GradNorm~\cite{chen2018gradnorm}, as opposed to fixing them throughout the training process~\cite{MTLfairWang0BPCC21}. 
Our contributions can be summarized as follows:
    i) We introduce the dynamic objective selection paradigm for fair and accurate MTL.
    ii) We propose a new algorithm, \our, based on a student-teacher framework that executes the dynamic objective selection paradigm and efficiently solves fairness-aware MTL. 
    iii) Our dynamic objective selection results in a reduction of parameters to be learned per training step from $2T$ to $T$.
    iv) We eliminate the dependency of searching for the correct balance of inter-task trade-off weights by automatically learning the weights at each training step.
   v) \our outperforms state-of-the-art methods by improving on both fairness and accuracy as demonstrated on real-world datasets of varying characteristics and number of tasks.

The rest of the paper is organized as follows: In Section~\ref{sec:related} we review the related work. Necessary basic concepts are provided in Section~\ref{sec:preliminaries}. Our method is introduced in  Section~\ref{sec:our}, followed by an experimental evaluation in Section~\ref{sec:experiments}. Conclusions and outlook are summarized in Section~\ref{sec:conclusion}. 

\section{Related work}
\label{sec:related}
\textbf{Fairness-aware learning}
A growing body of work has been proposed over the last years
to address the problem of fairness and algorithmic discrimination~\cite{mehrabi2021survey}
against demographic groups defined on the basis of protected attributes like gender or race. 
In parallel to bias mitigation methods, a plethora of  fairness notions have been
proposed; the interested reader is referred to~\cite{mehrabi2021survey} for a taxonomy of various fairness definitions. Statistical parity~\cite{dwork2012fairness}, equal opportunity and equalized odds~\cite{hardt2016equality} are among the most popular measures for measuring discrimination. In this work, we adopt equalized odds as our notion of fairness (Section~\ref{sec:fairdef}).


\noindent \textbf{Multi-task learning}
In MTL, multiple learning tasks are solved simultaneously, while exploiting commonalities and differences across the tasks~\cite{2021mtl_survey}. 
There are two main categories of parameter sharing: hard vs. soft. In hard parameter sharing~\cite{chen2018gradnorm,ecml_mtl_2021self,multi-task_gnn_2021deep,MTLfairWang0BPCC21}, 
model weights are shared between multiple tasks, while output layers are kept task-specific.
In soft parameter sharing~\cite{ecml_mtl_2021vogue}, different tasks have individual task-specific models with separate weights, but the distance between the model parameters is regularized in order to encourage the parameters to be similar.
In this work, 
 we follow the most popular hard parameter sharing  
approach where model weights are shared between multiple tasks.

\noindent \textbf{Fairness-aware multi-task learning} 
The fairness implications of MTL have been only recently considered: ~\cite{2019fair-multi-regression} studies multi-task regression to improve fairness in ranking;  
\cite{2019mtl-fair_cons} proposes an MTL formulation to solve multi-attribute fairness on a single task. 
The closest to our work is the seminal work~\cite{MTLfairWang0BPCC21}, which formulates the fair-MTL problem as a weighted sum of task-specific accuracy-fairness trade-offs. This formulation results in duplication of parameters, which are learned via hyperparameter tuning. Moreover, \cite{MTLfairWang0BPCC21} introduced the concept of task-exclusive labels signifying examples that are only positive (negative) for the concerned task and negative (positive) for all the other tasks.
Then, they proposed the \emph{MTA-F} algorithm that updates the task-specific layer with the summed loss of accuracy and task-specific fairness (computed with exclusive examples) and shared layers with the summed loss of accuracy and shared fairness  (computed with non-exclusive examples). In our method, we do not use the task-exclusive concept, as in the presence of a large number of tasks, the  exclusive set of instances may reduce to null. 
%
Moreover, contrary to~\cite{MTLfairWang0BPCC21} that assumes a fixed intra-task accuracy-fairness trade-off, we 
rather \emph{learn to choose} \emph{at each step} of the training process whether the accuracy loss or the fairness loss should be used for model training.  
Also, instead of fixing the task-specific weights, we propose to learn the right trade-off parameters at each step using GradNorm~\cite{chen2018gradnorm}.  

\section{Problem setting and basic concepts}
\label{sec:preliminaries}
We assume a set of tasks $T=\{1, \cdots, T\}$ sharing an input space 
$X=U\times S$, where $U$ is the subspace of \emph{non-protected attributes} and  $S$ is the subspace of \emph{protected attributes}.
Each task $t$ has its own label space $Y_t$.
A dataset $D$ of i.i.d. instances from the input space $X=U\times S$ and task spaces $\{Y_t\}_1^T$ is given as: $D=\{(u_i,s_i,y_i^1, \cdots y_i^T )\}_{i=1}^n$ where $(u_i,s_i)$ is the description of instance $i$ and $(y_i^1 \cdots y_i^T)$ are the associated class labels for tasks $1 \cdots T$.
For simplicity, we assume binary tasks, i.e., $\forall{t \in T}: Y_t\in \{1,0\}$, with $1$ representing a positive (e.g., ``granted") and $0$ representing a negative (e.g.,``rejected") class. 
We also assume the protected subspace $S$ to be a binary protected attribute:
$S\in \{g,\overline{g}\}$, where $g$ and $\overline{g}$ represent the \emph{protected} and \emph{non-protected group}, respectively. 

\subsection{Fairness Definition and Metric}
\label{sec:fairdef}
As our fairness measure, we use \verb|Equalized odds| ($EO$)~\cite{hardt2016equality}, introduced for STL.
For a task $t$, $EO$
states that a classifier's prediction $\hat{Y_t}$ conditioned on ground truth $Y_t$ must be independent from the protected attribute $S$. Based on~\cite{hardt2016equality}, fairness is preserved when: $P(\hat{Y_t}=1|S=g,Y_t=y)=P(\hat{Y_t}=1|S=\overline{g},Y_t=y)$, where $y\in\{1,0\}$. 
A classifier satisfies the  $EO$ definition 
if the protected and non-protected groups have equal true positive rate (TPR) and false positive rate (FPR). 
For a task $t$, the FPR w.r.t. the  protected group $g$ is given by:
$FPR_{t}(g)$: $P(\hat{Y_t}=1|Y_t=0,S=g)$, whereas the TPR is given by: 
$TPR_{t}(g)$: $P(\hat{Y_t}=1|Y_t=1,S=g)$. 
Similarly, we can define $FPR_{t}(\overline{g})$ and $TPR_{t}(\overline{g})$ for the non-protected group $\overline{g}$.
The (absolute) differences in TPR and FPR define the violation of $EO$ w.r.t.  $S$, denoted by $EO_{viol}$. Given $TPR=1-FNR$, the violation can be also expressed in terms of FNR and FPR differences:
\begin{equation}
\label{eq:eo_def}
      \begin{split}
    EO_{viol}=\Big|TPR_{t}(g)-TPR_{t}(\overline{g})\Big| +\Big|FPR_{t}(g)-FPR_{t}(\overline{g})\Big|\\
           = \Big|FNR_{t}(g)-FNR_{t}(\overline{g})\Big|+\Big|FPR_{t}(g)-FPR_{t}(\overline{g})\Big|
        \end{split}
    \end{equation}

\subsection{Vanilla Multi-task Learning (MTL)}\label{sec:mtl}
\begin{figure}
    \centering
    \includegraphics[width=\textwidth]{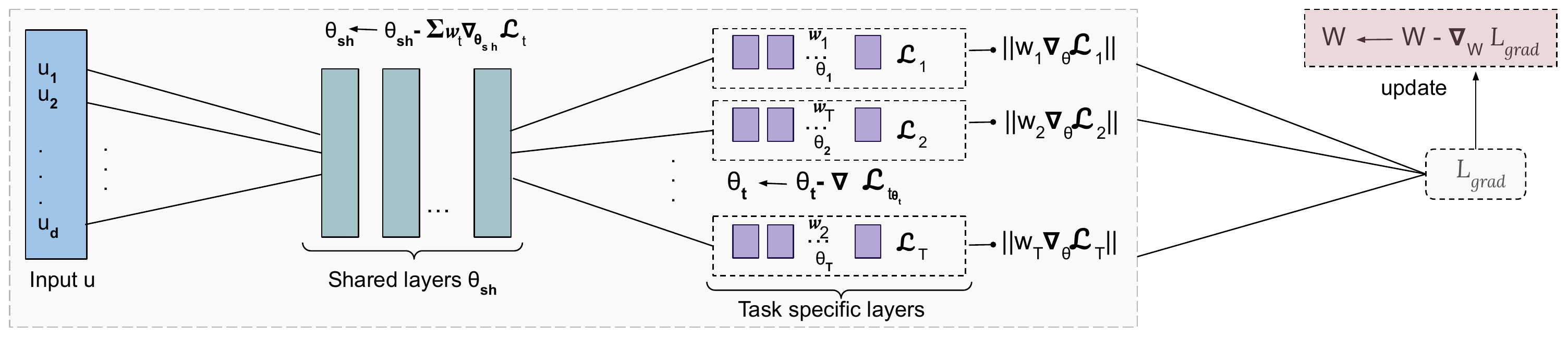}
    \caption{An overview of (vanilla) MTL with GradNorm update}
    \label{fig:gn_mtl}
\end{figure}

Let $\mathcal{M}$ be an MTL model
parameterized by the set of parameters $\theta \in \Theta$, which includes: shared parameters $\theta_{sh}$ (i.e., weights of layers shared by all tasks) and task-specific parameters $\theta_t$ (i.e., weights of the task specific layers), i.e.:   $\theta=\theta_{sh}\times \theta_1\times\cdots\times\theta_T$. An overview is given in Figure~\ref{fig:gn_mtl}.

The goal of vanilla MTL training is to jointly minimise multiple loss functions, one for each task:
$\argmin\limits_{\theta}(\mathcal{L}_1(\theta,U), \cdots, \mathcal{L}_T(\theta,U))$.
Finding a model $\theta$ that minimizes all $T$ tasks simultaneously is hard and typically, a scalarization approach is followed \cite{chen2018gradnorm,multi-task_gnn_2021deep,ecml_mtl_2021vogue}  which turns the multi-tasking function into a single function using task-specific weights $w_t$, as follows:
\begin{equation}\label{eq:mtl}    \argmin_\theta \sum_{t} w_t \mathcal{L}_t(\theta,U)
\end{equation}

The task-specific weights $w_t$ signify the importance of each task; 
usually selected by hyperparameter tuning over a validation set.  This, however, means \emph{fixing} the weights over the whole training process. 
A better approach, which we follow in this work, is to vary the weights at each training epoch to balance the contribution of each task for optimal model training. In this direction, GradNorm~\cite{chen2018gradnorm} proposes to balance the training rates of different tasks; if a task is training relatively quickly, then its weight should decrease relative to other tasks' weights to allow other tasks to influence training. GradNorm is implemented as an $L1$ loss function between the actual and target gradient norms at each training epoch for each task, summed over all tasks:
\begin{equation}\label{eq:grad_norm}
    L_{grad}=\sum_{t}|G_t(\theta,U) - \overline{G(\theta,U)}\times \rho_t^\alpha| \end{equation}
where $G_t(\theta,U)$ is the L2 norm of the gradient of the loss for task $t$  w.r.t. the chosen weights $\theta$, defined as $G_t(\theta,U)=||\nabla_\theta w_t\mathcal{L}_t(\theta,U)||_2$; $\overline{G(\theta,U)}$ is the average gradient norm.   
Finally,  $\rho_t$ is the inverse training rate for task $t$  and $\alpha$ is a hyperparameter of strength to pull any task back to the average training rate.

\subsection{Fairness-aware Multi-task Learning (FMTL)}\label{sec:mtlf}
Fairness has been extensively studied in the recent years for STL problems; most approaches combine accuracy and fairness losses into a single overall loss~\cite{padala2020fnnc} as:
\begin{equation}
\label{eq:fair_stl}
    \argmin_\theta \Big( \mathcal{L}(\theta,U)+\lambda \mathcal{F}(\theta,S)\Big)
\end{equation}
where $\mathcal{L}(\theta,U)$ is the typical \emph{accuracy loss}, $\mathcal{F}(\theta,S)$ is the loss associated with the protected attribute $S$ (refered to as \emph{fairness loss}) and $\lambda$ is a weight parameter that determines the fairness-accuracy trade-off.

Fairness-aware MTL extends traditional MTL and fair-STL setups by considering not only the predictive performance but also the fairness performance on the individual tasks. The fair-MTL problem was formulated~\cite{MTLfairWang0BPCC21} as optimizing a weighted sum of accuracy and fairness losses over all tasks:
\begin{equation}\label{eq:mtlfair_summed}
    \argmin_\theta \sum_t w_t \Big(\mathcal{L}_t(\theta,U)+\lambda_t \mathcal{F}_t(\theta,S)\Big)
\end{equation}

where $\lambda_t$  determines the fairness-accuracy trade-off within a task $t$ and $w_t$ determines the relative importance of task $t$. Note that with this formulation the number of parameters required to be learned is doubled from $T$ (for vanilla MTL, c.f., Eq.~\ref{eq:mtl}) to $2T$ (for FMTL c.f Eq.~\ref{eq:mtlfair_summed}).

\subsection{Deep Q-learning (DQN) and Multi-tasking DQN (MT-DQN)}\label{sec:dqn}
Reinforcement Learning (RL) is based on learning via interaction with a single environment.
At each step $j$, 
the environment observes a state $z_j \in Z$. The agent takes an action $a_j \in A$ in the environment, causing a transition to a new state $z_{j+1} \in Z$. For the transition, the agent receives a reward $R(j) 
\in \mathbb{R}$. 
The goal of the agent is to learn a policy $\pi: A\times Z \rightarrow A$ that maximizes the expected future discounted reward. 
State-action values (Q-values) are often used as an estimator of the expected future return. 
When the state-action space is large, the Q-values are typically approximated via a DNN, known as Deep Q-Network (DQN). The input to the DQN are the states, whereas the outputs are the state-action values (or, Q-values).
A DQN is trained  with parameters $\theta^Q$, to minimize
the loss between the \emph{predicted Q-values} and the   \emph{target future return}:
\begin{equation}\label{eq:DQN}
  L_Q(Z,A,\theta^Q,R)= E\Big[
   Q(z_j,\alpha_j|\theta^Q) -
   \Big( R(j)+\gamma \max_{\alpha_j'}Q(z_j',\alpha_j'|\theta^Q)\Big)\Big]
\end{equation}
where $Q(z_j,\alpha_j|\theta^Q)$ is the predicted Q-value.  
The term  $R(j)+\gamma \max\limits_{\alpha_j'}Q(z_j',\alpha_j'|\theta^Q))$ is the target value defined as the sum of the direct reward $R(j)$ for transitioning from state $z_j$ to a successor state $z'_j$ and of the Q-value of the best successor state~\cite{mnih2015human_dqn} (as predicted by the DQN). 

In \emph{multi-task reinforcement learning}, 
a single agent must
learn to master $T$ different environments/task, so the environments are the different tasks. 
The DQN for approximating the Q-values of state-action pairs is now a multi-tasking deep network (c.f., Section~\ref{sec:mtl}) with parameters 
$\theta^Q=\theta^Q_{sh}\times \theta^Q_1\times\cdots\times\theta^Q_T$, where $\theta^Q_{sh}$ are shared across all the task learning environments, and $\theta^Q_t$ are the task exclusive parameters for learning the optimal policy in the particular environment/task.  
The network should learn to predict the Q-values under different learning environments/tasks, so the new objective becomes:
\begin{equation}\label{eq:mtQ_opti}
   \argmin_{\theta^Q} \sum_t w^Q_t  L_Q(Z(t),A(t),\theta^Q, R(t))
\end{equation}
where  $w^Q_t$ is the learned weight for the environment/task $E_t$ to overcome the challenge of inter-task loss dominance.
\color{black}

\section{Learning to Teach Fairness-aware Multi-tasking}
\label{sec:our}

We first present the dynamic objective selection paradigm to formulate the fair-MTL problem (Section.~\ref{sec:dynamic}). Then we introduce our \our algorithm in Section~\ref{sec:algo}, which is a teacher-student network framework. The student network (Section~\ref{sec:studentNetwork}) learns a fair-MTL model, using advice from the teacher (Section~\ref{sec:teacherNetwork}) regarding the choice of the loss function.  

\subsection{Dynamic Loss Selection Formulation}\label{sec:dynamic}
The fair-MTL problem definition (cf. Eq.~\ref{eq:mtlfair_summed}) introduces the $\lambda_t$ parameters for combining accuracy and fairness losses for each task  and therefore, the number of parameters required to be learned is doubled from $T$ (for traditional MTL, cf. Eq.\ref{eq:mtl}) to $2T$ (for fairness-aware MTL, cf. Eq.~\ref{eq:mtlfair_summed})). 
We propose to get rid of the $\lambda$ dependency and rather \emph{learn to choose} \emph{at each training step} whether the accuracy loss ($\mathcal{L}_t$) or the fairness loss ($\mathcal{F}_t$) should be used to train the model on task $t$. Such an approach reduces the optimization problem of Eq.~\ref{eq:mtlfair_summed} to that of Eq.~\ref{eq:mtl}, which is well studied in the literature~\cite{chen2018gradnorm,multi-task_gnn_2021deep,ecml_mtl_2021vogue,2021mtl_survey}. 
Moreover, it also provides the flexibility to the learner to emphasise on the objective (accuracy or fairness) that is harder to learn for each task at each training step.

The transformed learning problem is formulated as:
\begin{equation}\label{eq:prob_form}
    \argmin_{\theta, m} \sum_t w_t (m) 
    \begin{cases}
    \mathcal{L}_t(\theta,U,m), \text{if task $t$ is trained for \emph{accuracy} in epoch $m$}\\
    \mathcal{F}_t(\theta,S,m), \text{if task $t$, is trained for \emph{fairness} in epoch $m$}
    \end{cases}
\end{equation}
where $m$ is the epoch\footnote{To note that the epoch information though specifically used in Eq.~\ref{eq:prob_form} to understand the temporal aspect of the selection, is valid for all the previous and following equations and is omitted for the rest of the paper for ease of reading.} number that sequences until model convergence. 

The decision about which loss to use is based on the teacher network, which is trained jointly with the student network (Section~\ref{sec:algo}).
The accuracy loss and fairness loss functions that we adopt in this work are provided hereafter.

\noindent \textbf{Accuracy loss}: We adopt the negative log-likelihood loss    $\mathcal{L}_t(\theta,U)$ for a task $t$: 
\begin{equation}\label{eq:task_error}
    \mathcal{L}_t(\theta,U)=-\sum_i ( y_{i}^t \log \mathcal{M}(u_{i},\theta) - (1-y_{i}^t) \log (1-\mathcal{M}(u_i,\theta)))
\end{equation}

\noindent \textbf{Fairness loss}: Several works discuss how to formulate a fairness loss function ~\cite{gretton2006kernel,feldman2015certifying,donini2018empirical,oneto2020general_empirical,rezaei2020robustlogloss}.
To keep the characteristic of the fairness loss function similar to the one used for the accuracy loss, we adopt
the robust log-loss~\cite{rezaei2020robustlogloss,padala2020fnnc},
 which focuses on the worst-case log loss and is given by:
\begin{equation}\label{eq:fair_error}
    \begin{split}
        \mathcal{F}_t(\theta,U,S)=\max(\mathcal{FNR}^{g}_{t}(\theta,U,S),\mathcal{FNR}^{\overline{g}}_{t}(\theta,U,S))\\
      +\max(\mathcal{FPR}^{g}_{t}(\theta,U,S),\mathcal{FPR}^{\overline{g}}_{t}(\theta,U,S))
    \end{split}
\end{equation}
where the negative log-likelihood loss over a group (say $g$) for a task $t$:
\begin{equation}\label{eq:fair_blocks}
    \begin{split}
      \mathcal{FNR}^{g}_{t}(\theta,U,S,Y_t)= -\sum_i y_{i}^t \log \mathcal{M}(u_{i},\theta|y_{i}^t=1,s_{i}=g)\\
       \mathcal{FPR}^{g}_{t}(\theta,U,S,Y_t)= -\sum_i (1-y_{i}^t) \log (1-\mathcal{M}(u_{i},\theta|y_{i}^t=0,s_{i}=g))
    \end{split}
\end{equation}



\begin{figure*}
    \centering
    \includegraphics[width=\textwidth]{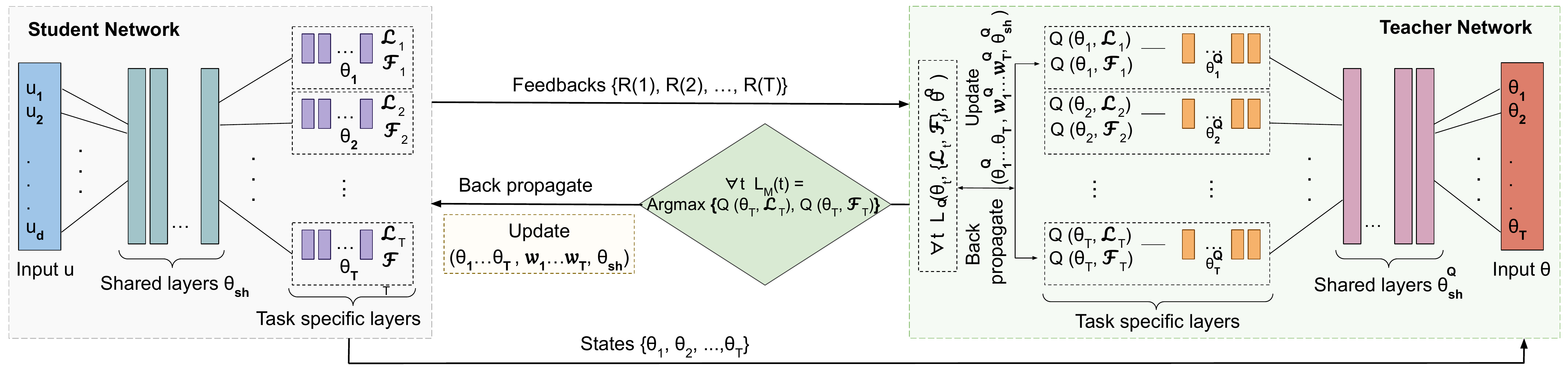}
    \caption{The L2T-FMT architecture. Steps are in order from bottom to top}
    \label{fig:l2t_arc}
\end{figure*}
\subsection{L2T-FMT Algorithm}\label{sec:algo}
An overview of our \emph{Learn to Teach Fairness-aware MTL (L2T-FMT)} method is shown in Figure~\ref{fig:l2t_arc}.
It consists of a student network and a teacher network which are trained collaboratively, as seen in Algorithm~\ref{algo:l2t}.

The \emph{teacher network}  $\mathcal{Q}$ aims at each training round to select (line 2) the best loss function for each task among the two available options: accuracy loss and fairness loss. The selected loss function is adopted by the student network $\mathcal{M}$ and is used for network update (line 3).
After the update, the \emph{student network} provides feedback (line 4) to the \emph{teacher network} about the progress made in each of the tasks following the advice of the \emph{teacher}.
 Based on the feedback, the \emph{teacher} network updates itself (line 5).
 So the two networks are trained collaboratively;  their functionalities are explained in the following sections. 

\begin{algorithm}[!htbp]
	\caption{The L2T-FMT algorithm}
	\label{algo:l2t}
		\KwInput{A MTL dataset $D\in U\times S \times Y$ (see Section~\ref{sec:preliminaries})}
	\KwInit{Initialize student model  $\mathcal{M}$  parameters: $\theta=\theta_{sh}\times \theta_1\times\cdots\times\theta_T$ and 
	task weights: $w_t=\frac{1}{T}$\\
	Initialise teacher model $\mathcal{Q}$: parameters- $\theta^Q=\theta^Q_{sh}\times \theta^Q_1\times\cdots\times\theta^Q_T$ and environment weights  $w^Q=\{w^Q_t=\frac{1}{T}\}$
	}
	\begin{algorithmic}[1]
	    \State \textbf{Until} convergence \textbf{do}
	    \State \phantom{abv} \small$d=\{\argmax\limits_{\alpha \in \{\mathcal{L}_t,\mathcal{F}_t\}}Q(\theta_t,\alpha), \forall t\}$\normalsize \tcp{teacher decides the best loss to learn for student upon seeing $\theta_t$}
        \State \phantom{abc} Call Algorithm~\ref{algo:stud} with input $(U,S,Y,d)$\tcp{Training $\mathcal{M}$}
        \State \phantom{abc} Evaluate $R=\{[\mathcal{R}(\mathcal{L}_t), \mathcal{R}(\mathcal{F}_t)]~\forall t\}$ using Eq.~\ref{eq:reward} \tcp{generate feedback for the teacher}
        \State \phantom{abc} Call Algorithm~\ref{algo:teach} with input $(\{\mathcal{M}(t)\}, \{\theta_t\},R, A)$ \tcp{Training $Q$}
	\end{algorithmic}
	\KwOutput{$\mathcal{M}$ with learned parameters $\theta=\theta_{sh}\times \theta_1\times\cdots\times\theta_T$}
\end{algorithm}

\subsection{Student Network}
\label{sec:studentNetwork}
\begin{algorithm}[!htbp]
	\caption{Student\_MTL}
	\label{algo:stud}
	\KwInput{ $U$, $S$, $Y=\{Y_t\}_1^T$, $d$ teacher's decision for the loss (line 2, Algo.~\ref{algo:l2t})}
	\begin{algorithmic}[1]
        \State \textbf{for} 1,\dots,T \textbf{do} 		
        \State \phantom{abc} \textbf{if} \small$d(t)=\mathcal{L}_t$ \normalsize\textbf{then} \small{$L_\mathcal{M}(t)=\mathcal{L}_t(\theta,U)$} \normalsize~per task (as in Eq.~\ref{eq:task_error}) \tcp{Compute accuracy loss }
        \State \phantom{abv} \textbf{else} \small{$L_\mathcal{M}(t)=\mathcal{F}_t(\theta,U,S)$}\normalsize~per task (as in Eq. \ref{eq:fair_error}) \tcp{Compute fairness loss}
		\State \phantom{abv} \small$\theta_t\leftarrow \theta_t - \eta \nabla_{\theta_t}L_\mathcal{M}(t)$\normalsize \tcp{Update  task-specific layers params}
		\State \phantom{abv} \small $G_t=||\nabla_\theta w_t L_\mathcal{M}(t)||_2$, and  $\rho_t=\frac{L_\mathcal{M}(t)}{E(L_\mathcal{M}(t))}$\normalsize \tcp{calculate gradient norm, and inverse training rate}
		\State \textbf{end for}
		\State Compute \small $L_{grad}$\normalsize~as in Eq.~\ref{eq:grad_norm} and update $W$ \small$\forall_{w_t\in W} w_t\leftarrow w_t- \eta \nabla_{w_t} L_{grad}$\normalsize 
		\State $\theta_{sh}\leftarrow \theta_{sh} - \eta \sum_t w_t\nabla_{\theta_{sh}}L_\mathcal{M}(t)$  \tcp*{update the shared parameters}
	\end{algorithmic}
\end{algorithm}
The student network $\mathcal{M}$ is a deep multi-tasking neural network with learning parameters $\theta=\theta_{sh}\times \theta_1\times\cdots\times\theta_T$, as described in Sec.~\ref{sec:mtl}. It aims to learn to solve the fairness-aware multi-tasking problem by optimizing Eq.~\ref{eq:prob_form}.
The pseudocode is shown in  Algorithm~\ref{algo:stud}.
For each task $t$, the decision of the \emph{teacher} $\mathcal{Q}$  (given as input) about which action/loss to use is followed. Based on the decision, the selected loss (accuracy or fairness) is  computed (lines 2-3) and used for the update of  task-specific network layers  (line 4). 
The task-specific weights $w_t$ are learned using GradNorm (lines 5 and 7). Finally, the shared parameters are updated (line 8) using the updated task-specific weights and the loss function decided by the teacher network. The weight update mechanism using GradNorm (c.f., Section~\ref{sec:mtl}) is visualized in Fig~\ref{fig:gn_mtl} for reference. 


\subsection{Teacher Network}
\label{sec:teacherNetwork}

The teacher network $\mathcal{Q}$ is a MT-DQN agent, as described in Sec.~\ref{sec:dqn}. It aims to learn to decide about which loss function, among the accuracy and fairness, the student network $\mathcal{M}$ should use. 
In particular, $\mathcal{Q}$ learns to predict the Q-values of accuracy-, fairness-loss/actions; the actual decision is the action with the largest Q-value (line 2 of Algorithm~\ref{algo:l2t}). 
 \begin{algorithm}[!htbp]
	\caption{Teacher\_DQN}
	\label{algo:teach}
	\KwInput{ $\{\mathcal{M}(t)\}$, $\{\theta_t\}$, $\{R(t)=[\mathcal{R}(\mathcal{L}_t),\mathcal{R}(\mathcal{F}_t)]\}$, $\{A(t)|A(t)=\{\mathcal{L}_t(\theta,U), \mathcal{F}_t(\theta,S)\}\}$, }
	\begin{algorithmic}[1]
        \State \textbf{for} 1,\dots,T \textbf{do} 
        \State \phantom{abc} Estimate $Q(\theta_t,\mathcal{L}_t)$, $Q(\theta_t,\mathcal{F}_t)$
        \normalsize~
        \tcp{estimating Q-values for each state-action pair}
		\State \phantom{abv} \small$\forall \alpha_t'\in A(t)$\normalsize~take action \small$\alpha_t'$ in $M(t)$\normalsize to produce \small$\theta_t'$\normalsize \tcp{teacher takes a one look forward into the student environment}
		\State\phantom{abv} Compute the loss  \small$L_Q(\theta_t,A(t),\theta^Q,R(t))$ \normalsize(~as in Eq.~\ref{eq:DQN})
		\State \phantom{abv}\small $\theta^Q_t \leftarrow \theta^Q_t - \eta \nabla_{\theta^Q_t}L_Q(\theta_t,A(t),\theta^Q)$\normalsize \tcp{teacher updates the parameters of the task environment specific layers}
		\State \phantom{abv} \small $G^Q_t=||\nabla_\theta w^Q_t L_Q(\theta_t,A(t),\theta^Q)||_2$, and  $\rho^Q_t=\frac{L_Q(\theta_t,A(t),\theta^Q)}{E(L_Q(\theta_t,A(t),\theta^Q))}$\normalsize \tcp{calculate gradients norm, and inverse training rate}
		\State \textbf{end for}
		\State Compute \small $L_{grad}^Q$\normalsize~as in Eq.~\ref{eq:grad_norm} and update $w^Q$ \small$\forall_{w^Q_t\in w^Q} w^Q_t\leftarrow w^Q_t- \eta \nabla_{w^Q_t} L_{grad}^Q$\normalsize
		\State $\theta^Q_{sh}\leftarrow \theta^Q_{sh} - \eta \sum_t w^Q_t\nabla_{\theta^Q_{sh}}L_Q(\theta_t,A(t),\theta^Q)$  \tcp{update the shared parameters}
	\end{algorithmic}
\end{algorithm}   

 The pseudocode of $\mathcal{Q}$ is shown in  Algorithm~\ref{algo:teach}. For each task $t$, it estimates the Q-values of the accuracy- and fairness-loss functions/actions based on the current model parameters (line 2). The network is updated (lines 3-9) as described in Section~\ref{sec:dqn}. 
Teacher's training depends on the feedback by the student network, in the form of reward,  
which is computed by evaluating the student's output in terms of $Acc$ (accuracy) and EO$_{viol}$ (fairness) (line 4, Algorithm~\ref{algo:l2t}).  
The main intuition for the design of the reward function is to reward positively only if $Acc$ increases and simultaneously $EO_{viol}$ decreases for any action suggested by $Q$ for a task $t$. On violation of either of the two conditions, the reward should be negative (the min function ensures this positive/negative property). 
However, 
the problem of intra-task dominance arises in estimation of $\mathcal{R}$ as the scales of evaluated \textit{accuracy} and \textit{fairness} for task $t$ might differ. To calculate a scale-invariant reward, we take inspiration from~\cite{hessel2019multiRLpopart}, and estimate $\mathcal{R}(\alpha_t)\in \mathbb{R}$ for $\alpha_t \in \{\mathcal{L}_t,\mathcal{F}_t\}$,  as the transformed evaluated output:
\small\begin{equation}
    \begin{split}\label{eq:reward}
       \mathcal{R}(\alpha_t)=\min(\frac{Acc(\hat{Y}_t)-Acc^{best}(t)}{Acc^{best}(t)},\frac{EO_{viol}^{best}(t) - EO_{viol}(\hat{Y}_t)}{1-EO_{viol}^{best}(t)})
    \end{split}
\end{equation}\normalsize
where $Acc(\hat{Y}_t)$, $Acc^{best}(t)$, and $EO_{viol}(\hat{Y}_t)$, $EO_{viol}^{best}(t)$ are respectively the current and best till current epoch \textit{accuracy}, and \textit{fairness} values of $\mathcal{M}$ in task $t$. 


\section{Experiments}
\label{sec:experiments}
 We first evaluate the accuracy-and fairness of our \our\footnote{https://anonymous.4open.science/r/L2TFMT-F309/} in comparison to other approaches for different MTL problems (Section~\ref{sec:expOveralEvaluation}) including a more task-specific evaluation (Section~\ref{sec:expPerTask}).
 Next, we analyse the impact of the dynamic loss function selection by \our~(Section~\ref{sec:exp_LossEffect}). 
The experimental setup including datasets, evaluation measures, and competitors 
is discussed in Section~\ref{sec:expSetup}. 

\subsection{Experimental Setup}
\label{sec:expSetup}
\label{sec:expDatasets} \textbf{Datasets}
We evaluate on one tabular and two visual datasets.
The tabular dataset is the recently released  \textit{ACS-PUMS}~\cite{ding2021retiring}, which comprises a superset of the popular Adult dataset from  available US Census sources,  and consists of 5 different well defined 
binary classification tasks\footnote{\url{https://github.com/zykls/folktables}}. 
We use \textit{gender} as the protected attribute. For training we use the census data from the year ``2018", divided into train (70\%) and validation (30\%) sets. For testing we use the data from the following year ``2019" (both years of size $\approx 1.65M$). The visual datasets come from CelebA dataset~\cite{liu2015celebA} consisting of $202.5K$ celebrity facial images and 40 different binary attributes. We use the provided\footnote{\url{http://mmlab.ie.cuhk.edu.hk/projects/CelebA.html}} partitioning into train (\#162,770 instances), validation (\#19,867 instances), and test (\#19,962 instances) set.
We use two different protected attributes, gender and race, resulting into two versions of the dataset, \textit{CelebA-Gender} and \textit{CelebA-Race}. 
We don't consider all 40 attributes as tasks for the MTL since some attributes are extremely skewed towards the protected or non-protected group. For example, the attribute
“Mustache” is true only for 3 female instances. 
We set the filtering threshold to 1.5\% or 2.5K instances. 
The filtering process reduces the number of attributes to 17 for \textit{CelebA-Gender} and 31 for \textit{CelebA-Race}; these are the MTL tasks.
Further details on the datasets and experimental setup are provided in the Appendix.

\noindent \textbf{Methods}
We compared \our~against the following methods: 

\noindent \textit{i) MTA-F:}
the vanilla fairness-aware MTL method~\cite{MTLfairWang0BPCC21} that minimises the weighted sum of accuracy- and fairness-losses (c.f. Eq.~\ref{eq:mtlfair_summed}). For fairness it calculates two separate loss, one for updating $\theta_t$ and another for updating $\theta_{sh}$. 
The 
weights $w_t$ and 
$\lambda_t$ are set via hyperparameter tuning.

\noindent \textit{ii) G-FMT:} a variation of our \our~approach that always chooses \emph{greedily} the best action/loss function among the available choices, by optimising:
\small${argmin}_\theta \sum_t w_t \{\max \{\mathcal{L}_t(\theta,U), \mathcal{F}_t(\theta,S)\} \}$ \normalsize.

\noindent \textit{iii) Vanilla MTL:} the vanilla MTL approach that does not consider fairness but aims at minimising the weighted sum of task-specific accuracy losses (c.f. Eq.~\ref{eq:mtl}). The task-specific weights $w_t$ are learned via GradNorm~\cite{chen2018gradnorm} as in Eq~\ref{eq:grad_norm}.

\noindent \textit{iv) STL:} 
trains a separate fair-accurate model on each respective task.


\noindent \textbf{Evaluation Measures}\label{sec:evalMeasures}
Following~\cite{MTLfairWang0BPCC21}, we report on the relative performance of the \verb|MTL| model ($Acc(t)_{mtl}$, $EO_{viol}(t)_{mtl}$) to the performance of a \verb|STL| model trained on each respective task $t$ ($Acc(t)_{stl}$, $EO_{viol}(t)_{stl}$). 
Specifically, for accuracy we report on the \emph{average relative Acc} $(ARA)$:  
\small$ARA=\frac{1}{T}\sum_t^T\frac{Acc(t)_{mtl}}{Acc(t)_{stl}}$\normalsize
and for fairness, on the \emph{average relative EO} $(AREO)$: 
\small$AREO=\frac{1}{T}\sum_t^T\frac{EO_{viol}(t)_{mtl}}{EO_{viol}(t)_{stl}}$.
\normalsize 



\subsection{Overall fairness-accuracy evaluation}
\label{sec:expOveralEvaluation}

The overall fairness (AREO) and accuracy (ARA) performance of the different methods on all the datasets is shown in Table~\ref{tab:exp}.
As we can see, \our~outperforms all the competitors in \emph{fairness} by producing the lowest AREO scores across all the datasets; the relative reduction in discrimination 
w.r.t. the best baseline is in the range $[12\%-19\%]$.
Interestingly, the second best approach in terms of AREO is our greedy variation, \emph{G-MFT}. 
In terms of ARA, \our~is best by a small margin comparing to the best baseline with the exception of the ACS-PUMPS dataset for which Vanilla-MTL scores first.
In particular, for \textit{CelebA-Gender} and \textit{CelebA-Age}, our \our~beats the best baseline by $2\%$ and 1\%, respectively, whereas for the \textit{ACS-PUMPS} dataset \our scores second with a $3\%$ decrease comparing to the best performing \emph{Vanilla MTL}.

To get better insights on the results, in the next section
we also report on the task-specific performance using accuracy and $EO_{viol}$ for each task.

\begin{table}[]
\caption{ARA vs AREO:
Higher values better for accuracy/ARA, lower values better for discrimination/AREO . Best values in bold, second best underlined. $(\%)$ indicates our relative difference over the performance of the best baseline.}
\label{tab:exp}
    \centering
    \begin{tabular}{cclcccc}
        \hline
\multicolumn{1}{l|}
{\begin{tabular}[c]{@{}c@{}}\textbf{Dataset}\end{tabular}} &
  \multicolumn{1}{l|}{\textbf{\#tasks} $T$} &
   \multicolumn{1}{l|}{\textbf{Metric}} &
  \multicolumn{1}{l|}{\textbf{Vanilla MTL}} &
  \multicolumn{1}{l|}{\textbf{MTA-F}} &
  \multicolumn{1}{l|}{\textbf{G-FMT}} &
  \multicolumn{1}{l}{\textbf{L2T-FMT}} \\ \hline
\multirow{2}{*}{\begin{tabular}[c]{@{}c@{}}\textbf{ACS-}\\\textbf{PUMS}\end{tabular}} & \multirow{2}{*}{5}  & ARA  & \textbf{1.06} & 0.97 & 1.01      & \underline{1.03 (-3\%)}          \\ 
&                     & AREO & 2.38          & 3.52 & \underline{1.50}      & \textbf{1.21 (-19\%)} \\ \hline
\multirow{2}{*}{\begin{tabular}[c]{@{}c@{}}\textbf{CelebA-}\\\textbf{Gender}\end{tabular}}    & \multirow{2}{*}{17} & ARA  & 0.89          & \underline{0.95} & 0.86      & \textbf{0.97 (2\%)} \\ 
&                     & AREO & 2.72          & 2.29 & \underline{1.77}      & \textbf{1.51 (-15\%)} \\ \hline
\multirow{2}{*}{\begin{tabular}[c]{@{}c@{}}\textbf{CelebA}-\\\textbf{Age}\end{tabular}}       & \multirow{2}{*}{31}  & ARA  &    \underline{0.94}           &   0.85   &    0.86       & \textbf{0.95 (1\%)}     \\ 
&                     & AREO &      2.61         & 1.79     & \underline{1.72}  &     \textbf{1.52 (-12\%)}          \\ \hline
    \end{tabular}
\end{table}


\subsection{Performance distribution over the tasks}
\label{sec:expPerTask}
\begin{figure}[httb]
    \centering
    \begin{subfigure}[b]{0.4\columnwidth}
         \centering
         \includegraphics[width=\linewidth]{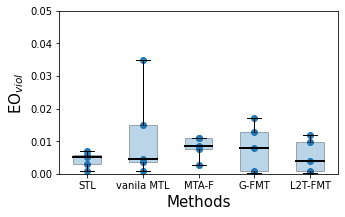}
         \caption{Boxplot fairness}\label{fig:box_pums_fair}
    \end{subfigure}
    \begin{subfigure}[b]{0.39\columnwidth}
         \centering
         \includegraphics[width=\linewidth]{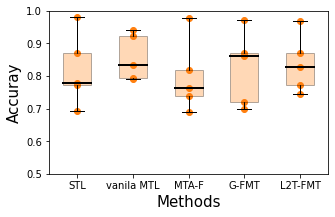}
         \caption{Boxplot accuracy}\label{fig:box_pums_acc}
    \end{subfigure}
    \caption{ACS-PUMS dataset: Performance distribution over the tasks.}
    \label{fig:boxplot_pums}
\end{figure}

\begin{figure}
    \centering
    \begin{subfigure}[b]{0.4\columnwidth}
         \centering
         \includegraphics[width=\linewidth]{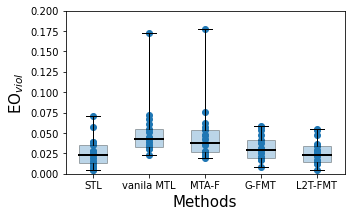}
         \caption{Boxplot fairness}\label{fig:box_gen_fair}
    \end{subfigure}
    \begin{subfigure}[b]{0.39\columnwidth}
         \centering
         \includegraphics[width=\linewidth]{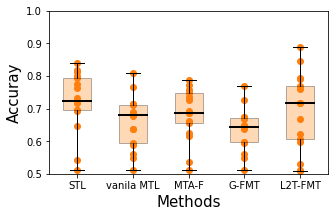}
         \caption{Boxplot accuracy}\label{fig:box_gen_acc}
    \end{subfigure}
    \caption{CelebA-Gender: Performance distribution over the tasks.}
    \label{fig:boxplot_gen}
\end{figure}

\begin{figure}[httb]
    \centering
    \begin{subfigure}[b]{0.4\columnwidth}
         \centering
         \includegraphics[width=\linewidth]{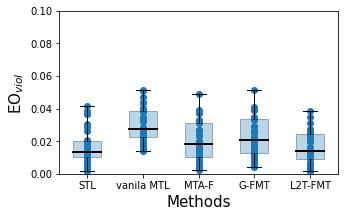}
         \caption{Boxplot fairness}\label{fig:box_age_fair}
    \end{subfigure}
    \begin{subfigure}[b]{0.39\columnwidth}
         \centering
         \includegraphics[width=\linewidth]{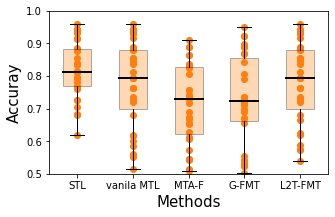}
         \caption{Boxplot accuracy}\label{fig:box_age_acc}
    \end{subfigure}
    \caption{CelebA-Age: Performance distribution over the tasks.}
    \label{fig:boxplot_age}
\end{figure}

We analyze the distribution of accuracy and fairness scores over the tasks for all methods using boxplots. The results are shown in Fig.~\ref{fig:boxplot_pums}, ~\ref{fig:boxplot_gen}, and ~\ref{fig:boxplot_age} for the \textit{ACS-PUMS}, \textit{CelebA-Gender} and \textit{CelebA-Age}, respectively. 
%
For fairness a positively skewed box (median closer to Q1) with low Q1 is better, while for accuracy a negatively skewed box (median closer to Q3) with high Q3 is better. 

\noindent
\textbf{ACS-PUMS dataset.}
In Fig~\ref{fig:box_pums_fair}, we see that \our~has the lowest \textit{median}, and the lowest \textit{Q1} of EO$_{viol}$, with \textit{Q3} marginally above the \textit{STLs} but lower than all the competitors. Henceforth, it achieves the best AREO score (c.f., Table~\ref{tab:exp}).
In Fig~\ref{fig:box_pums_acc}, we see that \our~has the second highest median after G-FMT, however it has a higher \textit{Q1} than \textit{G-FMT} but a lower \textit{Q1} and lower \textit{Q3} than \textit{vanilla MTL}. Thus, in Table~\ref{tab:exp} we find \our~to be second best in ARA score behind \textit{vanilla MTL} on ACS-PUMS.
 \textit{MTA-F} has the most consistent outcome of EO$_{viol}$ with low spread over the tasks for both fairness and accuracy, but has the highest median and \textit{Q1} of EO$_{viol}$, and the lowest median and lowest \textit{Q3} of accuracy. 
  Thus, in overall it has the worst overall performance as also seen in Table~\ref{tab:exp}.
 \textit{G-FMT} has the highest spread in both fairness and accuracy and is negatively skewed in accuracy with high accuracy for some tasks.
 \textit{Vanilla-MTL} comes second in terms of spread and is positively skewed in EO$_{viol}$. 
 However, 
 its upper whisker is longer with a single point above Q3; this corresponds to task 1 (\textit{Employment Status}) for which the EO$_{viol}$ score is high. 
 
\noindent
\textbf{CelebA-Gender dataset.}
In Fig~\ref{fig:box_gen_fair}, we see that all methods have low spread indicating their consistent performance w.r.t. fairness across the tasks. 
Still, \our~outperforms the competitors with the lowest median, \textit{Q1}, and \textit{Q3} values. \textit{MTA-F} has the third best median, \textit{Q1}, and \textit{Q3} of EO$_{viol}$,. As we see in Fig~\ref{fig:box_gen_fair} it has a much longer upper whisker indicating larger variance among the larger values, i.e., tasks with higher discrimination with the worst discrimination of 0.175 which corresponds to task 11 (\textit{Narrow Eyes}).  
However, in accuracy all the methods vary substantially (high spread) as we see in Fig~\ref{fig:box_gen_acc}. 
\our~holds the highest median and \textit{Q3}, but 
its \textit{Q1} is marginally lower than \textit{MTA-F} which indicates that in a few tasks \our~gets outperformed by \textit{MTA-F} (second best median, \textit{Q3}) on accuracy. This explains the ARA scores in Table~\ref{tab:exp}, where  \our~scores first followed by \textit{MTA-F}. 
\color{black}
\color{black}
\textit{G-FMT} has the second best median, \textit{Q3}, and \textit{Q1} of EO$_{viol}$ score, but has the worst median, \textit{Q3}, and \textit{Q1} of accuracy. Thus, in Table~\ref{tab:exp} we see that \textit{G-FMT} bags the second best AREO score but has the worst ARA score.

\noindent \textbf{CelebA-Age dataset.}
There is large spread 
accuracy (Fig~\ref{fig:box_age_acc}) across the different methods.
\our~has the lowest median, \textit{Q3}, and \textit{Q1} of EO$_{viol}$ score, and highest median, \textit{Q3}, and \textit{Q1} of accuracy, respectively. This reflects in Table~\ref{tab:exp} where \our~achieves the best AREO, and ARA scores.
\textit{MTA-F} delivers consistent fair performance over the tasks with the second best median, \textit{Q3}, and \textit{Q1} of EO$_{viol}$ score almost same as \textit{G-FMT}. This is the reason why the AREO score of \textit{G-FMT} and \textit{MTA-F} are nearly same, with \textit{G-FMT} marginally ahead, making \textit{MTA-F} the third best on AREO score. On accuracy \textit{G-FMT} has the third best median, \textit{Q3}, and \textit{Q1}, having marginal improvements over \textit{MTA-F}.
\textit{Vanilla MTL} with no fairness treatment has the worst median, \textit{Q3}, and \textit{Q1} of of EO$_{viol}$ score over the tasks positioning it at the last place on AREO score, while having median, \textit{Q3}, and \textit{Q1} of accuracy over the tasks very similar to that of \our~making it a very close second on ARA score.

\color{black}


\noindent
\textbf{Summary.} For all datasets, \our~stands out among the competitors with a very low median of EO$_{viol}$, and a very high median of accuracy. 
The fairness performance of \textit{MTA-F} depends on the number of tasks; for larger MTL problems (like \emph{CelebA-Gender and CelebA-Race})  the performance varies across the tasks including tasks with high discrimination (high upper whisker).
\textit{G-FMT} consistently delivers descent fairness performance positioning it always in the second place, however its accuracy gets affected when the number of tasks is high (CelebA-Gender and CelebA-Age). The \textit{Vanilla MTL} without any fairness treatment on EO$_{viol}$ score is often bad with very high upper whiskers. 


\subsection{Dynamic loss selection}
\label{sec:exp_LossEffect}

\begin{figure}[!htbp]
    \centering
    \begin{subfigure}[b]{0.17\columnwidth}
         \centering
         \includegraphics[width=\linewidth]{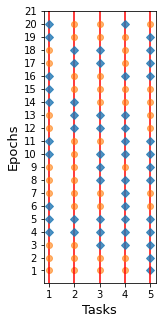}
         \caption{\tiny{ACS-PUMS}}\label{fig:pointers_pums}
    \end{subfigure}
    \begin{subfigure}[b]{0.34\columnwidth}
         \centering
         \includegraphics[width=\linewidth]{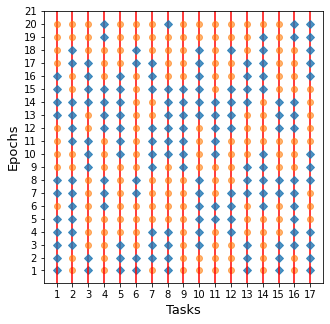}
         \caption{CelebA-Gender}\label{fig:pointers_gen}
    \end{subfigure}
    \begin{subfigure}[b]{0.46\columnwidth}
         \centering
         \includegraphics[width=\linewidth]{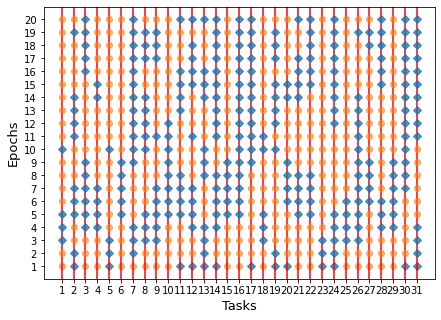}
         \caption{CelebA-Age}\label{fig:pointers_age}
    \end{subfigure}
    \caption{Loss function selection for each task over the training epochs. Accuracy-loss ($\mathcal{L}$) selection in orange, fairness-loss ($\mathcal{F}$) selection in blue.
}\label{fig:loss_point}
\end{figure}

We focus on the (dynamic) loss selection of the teacher network by looking at which function among the two available options: accuracy loss ($\mathcal{L}$) and fairness loss ($\mathcal{F}$) is used for each task over the training process.
The results for the different datasets are shown in Figure~\ref{fig:loss_point}. 

Regarding \emph{ACS-PUMS} (Fig.~\ref{fig:pointers_pums}), the decision of selecting $\mathcal{L}$ or $\mathcal{F}$ is almost equally distributed 
over the tasks. Using \emph{Vanilla MTL} (trained only for accuracy) as a reference method, we see in Table~\ref{tab:exp} that to achieve the best accuracy (best ARA) in this dataset one needs to always tune with $\mathcal{L}$. However, this comes with depreciation in fairness (high AREO). Thus, the necessary balance as selected by \our~is required. 

For \emph{CelebA-Gender} (Fig.~\ref{fig:pointers_gen}), the accuracy loss $\mathcal{L}$ is selected more often in some tasks (2, 10, 17) and the fairness loss  $\mathcal{F}$ in other tasks (6, 9, 11, 15), whereas there are also tasks with balanced selection (e.g.,  1).
\emph{Vanilla MTL} that only tunes for $\mathcal{L}$ does not produce the best accuracy (c.f., Table~\ref{tab:exp}); in contrast, \our~with dynamic loss selection achieves the best accuracy and fairness, justifying the selection (im-) balance.

For \emph{CelebA-Age}, (Fig.~\ref{fig:pointers_age}) we notice that for the majority of the tasks (21 out of 31) the loss selection is unevenly distributed with 13 tasks in favour of $\mathcal{L}$, and 8 in favour of $\mathcal{F}$. 
Interestingly, in 9 tasks (1, 5, 6, 10, 15, 18, 23, 25, 29) $\mathcal{L}$ is selected continuously over epochs at a stretch ($\geq 7$), signifying that in these tasks tuning for accuracy is far more important. This is also reflected in the very close ARA performance of \our and \emph{Vanilla MTL} (c.f., Table~\ref{tab:exp}).
Similarly, in 8 tasks (7, 14, 16, 17, 24, 26, 30, 31) $\mathcal{F}$ is chosen more frequently, which ultimately leads to the superiority of \our~in the AREO score. 

\color{black}


\section{Conclusion}\label{sec:conclusion}
We proposed \our, an approach for fairness-aware multi-task learning that dynamically selects for each task the best loss function to be used at each training step among the available: accuracy loss and fairness loss. Our approach is a student-teacher network framework, where the student learns to solve the fair-MTL problem while the teacher decides the action/loss function that the student should use for its update. The teacher is implemented as a DQN, whereas the student is implemented as a deep MTL network. 
In contrast to a rigid fairness-accuracy trade-off formulation~\cite{MTLfairWang0BPCC21}, \our~allows for a flexible model update based on which objective (accuracy or fairness) is harder to learn for each task. Moreover, it reduces the number of parameters to be learned to half. 
Our experiments on three real datasets show that \our~improves on both fairness and accuracy over state-of-the-art approaches. Moreover, we also show the effectiveness of learning to select the best loss in producing such favourable outcomes.



As part of our future work, instead of jointly training on all the tasks, we plan to identify the tasks that would benefit from training together. Such task groupings might differ based on whether fairness or accuracy is considered.


\bibliographystyle{splncs04}
\bibliography{biblio}


\end{document}